\documentclass[sigconf]{acmart}
\AtBeginDocument{%
  }

\setcopyright{acmlicensed}
\copyrightyear{2025}
\acmYear{2025}
\acmConference[ICAIL]{International Conference on Artificial Intelligence and Law}{June 16--20,
  2025}{Chicago, IL}

\patchcmd{\quote}{\rightmargin}{\leftmargin 2.3em \rightmargin}{}{}

\begin{document}

\title{Logical Modalities within the European AI Act: An Analysis}


\author{Lara Lawniczak}
\email{lara.lawniczak@uni-bamberg.de}
\orcid{1234-5678-9012}
\affiliation{%
  \institution{University of Bamberg}
  \city{Bamberg}
  \state{Bavaria}
  \country{Germany}
}

\author{Christoph Benzmüller}
\email{christoph.benzmueller@uni-bamberg.de}
\orcid{0000-0002-3392-3093}
\affiliation{%
  \institution{University of Bamberg}
  \city{Bamberg}
  \state{Bavaria}
  \country{Germany\\}
\institution{Freie Universität Berlin, Berlin, Germany}
}




\renewcommand{\shortauthors}{}

\begin{abstract}
    The paper presents a comprehensive analysis of the European AI Act in terms of its logical modalities, with the aim of preparing its formal representation, for example, within the logic-pluralistic Knowledge Engineering Framework and Methodology (LogiKEy). LogiKEy develops computational tools for normative reasoning based on formal methods, employing Higher-Order Logic (HOL) as a unifying meta-logic to integrate diverse logics through shallow semantic embeddings. This integration is facilitated by Isabelle/HOL, a proof assistant tool equipped with several automated theorem provers. The modalities within the AI Act and the logics suitable for their representation are discussed. For a selection of these logics, embeddings in HOL are created, which are then used to encode sample paragraphs. Initial experiments evaluate the suitability of these embeddings for automated reasoning, and highlight key challenges on the way to more robust reasoning capabilities.
\end{abstract}


  \label{fig:teaser}
\received{27 January 2025}
\received[revised]{12 March 2025}
\received[accepted]{5 June 2025}

\maketitle

\section{Introduction}
\indent In the last decade, the representation of, and the reasoning with, legal information has gained growing attention in the AI and Law community. Alongside established approaches like Akoma Ntoso \cite{akoma_ntoso_oasis} and LegalRuleML \cite{legalRuleML-1, legalRuleML-2},  research is currently being invested in the exploration and application of the logic-pluralistic knowledge engineering framework and methodology LogiKEy \cite{benzmuller-logikey}, which develops computational tools for normative reasoning based on formal methods.

An interesting and relevant use case for the LogiKEy framework is the European AI Act \cite{ai-act}, which came into force in August 2024. The AI Act classifies AI systems into three risk levels, and assigns different rules and regulations to each category. If successful, the logical formalisation of key concepts of the AI Act within the LogiKEy framework could, for example, prepare the way for sophisticated automated or interactive compliance checks and support in various ways the technical enforcement of the new regulation. It is not only the AI Act itself that is of interest, but also the specific follow-up regulations and standardisations that will be triggered by the AI Act, which will then impose specific regulatory constraints on concrete applications of AI systems.

Before adequate logical representations can be developed, it is essential to identify the modalities that occur in the AI Act and the logics, or combinations of logics, suited to adequately represent these. This paper addresses this challenge as its main contribution. In the long run, however, it is rather the concrete \textit{instantions} of the abstract legislation of the AI Act, i.e.~the more concrete laws that are expected to follow from the adoption of the AI Act in the following years, that are relevant and interesting for formalisation. We expect that the findings on the identified modalities will (to a large extent) overlap and apply to them as well.

The paper is organized as follows. Section 2 lists and discusses the modalities found in the relevant parts of the AI Act. The findings of this section are relevant beyond the LogiKEy context. Section 3 explores potential logic systems for representing these modalities. Section 4 briefly presents preliminary results on embedding selected logics in the LogiKEy framework and using them to formalize some exemplary sections of the AI Act. Section~5 concludes the paper.

\section{Modalities in the AI Act}
This section presents the modalities (and other challenges) identified in the AI Act and illustrates them with examples. In order to find all the modalities in the document, it was read several times, with different foci of interest, and visualized with the help of mind maps and tables.

\subsection{Obligations, Prohibitions, and Permissions}
The AI Act is a European piece of legislation intended to regulate the use of AI systems.
In it \textit{Obligations} are expressed with the word "shall", \footnote{Within legal acts such as the AI Act, the world "shall" can also have performative or constitutive functions instead of deontic ones. However, the primary function of "shall" in the AI Act, is the introduction of deontic obligations.} as in this sentence from Article 8: 
\begin{quote}
“High-risk AI systems shall comply with the requirements established in this section.” \cite[\S8(1)]{ai-act}
\end{quote}     
The negated shall, "shall not", is employed to express \textit{Prohibitions}. An example is: 
\begin{quote}
"Where an importer has sufficient reason to consider that a high-risk AI system is not in conformity with this Regulation, or is falsified, or accompanied by falsified documentation, it shall not place the system on the market until it has been brought into conformity." \cite[\S23(2)]{ai-act}
\end{quote}
For the expression of \textit{Permissions}, two strategies can be identified in the AI Act: They are introduced by either "may" or "is/are empowered to".

\subsection{Contrary-to-Duty-Obligations}

A special type of obligations that occurs within the AI Act deserves attention: \textit{contrary-to-duty obligations} (CTDs) are obligations that arise only if a primary obligation is not fulfilled, meaning that it is "conditional on [...] violating [a] primary obligation" \cite{sep-logic-deontic}.

An example of a CTD in the AI Act can be found in Article 20: 
\begin{quote}
"Providers of high-risk AI systems which consider or have reason to consider that a high-risk AI system that they have placed on the market or put into service is not in conformity with this Regulation shall immediately take the necessary corrective actions to bring that system into conformity, to withdraw it, to disable it, or to recall it, as appropriate." \cite[\S20(1)]{ai-act}
\end{quote}

It has been stated before that high-risk systems must comply with the requirements \cite[\S16(1)]{ai-act}. This is the primary obligation in the given context. However, obligations are not always fulfilled; they may also be violated. In the event of such a violation, the obligation to take corrective action becomes relevant. This is a CTD, applicable only when the primary obligation has been violated. CTD situations are often represented in a typical structure following \citeauthor{Chisholm} \cite{Chisholm}. The discussed example then looks as follows:
\begin{quote}
\begin{enumerate}
\item It ought to be that high-risk AI systems comply with the requirements in the regulation.
\item It ought to be that if a high-risk AI system does not comply with the requirements, providers take corrective actions.
\item If a system complies with the requirements, the provider must not take any corrective actions. \footnote{It has been brought to our attention that an interpretation as "need not" might be more appropriate. We are thankful for this remark and aim to uncover the differences between these interpretations in the future.}
\item Concrete Situation: The system does not comply with the requirements.
\end{enumerate}
\end{quote}

\indent Notice that some of these obligations in this example are agentive for the provider. The aspects of agency and agentive obligations are discussed separately in Section 3. First, two special kinds of CTDs are considered.

\subsubsection{Contrary-to-Duty-Obligations Involving Multiple Agents}
Some CTDs within the AI Act relate to multiple agents,  which complicates matters, as illustrated by the following example from $\S36$: 

\begin{quote}
"If the notifying authority comes to the conclusion that the notified body no longer meets the requirements laid down in Article 31 or that it is failing to fulfil its obligations, it shall restrict, suspend or withdraw the designation as appropriate, depending on the seriousness of the failure to meet those requirements or fulfil those obligations." \cite[\S36(4)]{ai-act}
\end{quote}

For simplicity, the notion of an agent belief, expressed in the words \textit{comes to the conclusion that}, is ignored (since beliefs are the subject of Section 4). Also, disregard the temporal notion contained in \textit{no longer}, since CTDs involving temporality are discussed in Section 2.2.2.

The focus here lies on the involvement of two distinct agents: The notifying authority and the notified body. The situation is as follows: The notified body has several primary obligations that are specified within the AI Act~\cite[\S31]{ai-act}. If it violates these obligations, another obligation arises. However, this obligation is agentive for the notifying authority, not for the notified body that violated the primary obligation.

\subsubsection{Contrary-to-Duty-Obligations Involving Temporality}

There are cases in the AI Act where CTDs occur in combination with temporal constraints, stating that a certain property used to be fulfilled but \textit{no longer} is. An example is:

\begin{quote}
"Where a notifying authority has sufficient reason to consider that a notified body no longer meets the requirements laid down in Article 31, or that it is failing to fulfil its obligations, the notifying authority shall without delay investigate the matter with the utmost diligence." \cite[\S36(4)]{ai-act}
\end{quote}

To understand the CTD, $\S30$ must be taken into account. There it is stated that the notifying authority "may notify only conformity assessment bodies which have satisfied the requirements laid down in Article 31" \cite[\S30(1)]{ai-act}. 
Unfortunately, this example does not  involve only temporality but also multiple agents as discussed in the Section 2.2.1. To focus on only temporality, we reformulate the regulation as follows: 

\begin{quote}
Only conformity assessment bodies that fulfill the requirements in Article 31 before the notification may be notified (adapted from Article 30). If a conformity assessment body that was notified (= notified body) no longer meets the requirements laid down in Article 31, the matter shall be investigated further with the utmost diligence (adapted from Article 36).
\end{quote}

Now it becomes visible how this example is different from a typical CTD: Technically speaking, the notified body is only obligated to fulfill the requirements in Article 31 at one point in time \textit{before} being notified. Hence, the fact that a notified body \textit{no longer } fulfills the requirements in Article 31 does not equal a violation of the previous obligation. The obligation to further investigate the matter then is just a normal obligation, not a CTD, and could be expressed as such. In this case, the logic representing this situation must not only provide adequate obligation operators, but must also be able to express temporality in order to properly capture the notion of \textit{before} (as discussed in Section 2.5). 

Another possibility is disregarding the temporality here and translating the example to match the usual CTD structure:
\begin{quote}
\begin{enumerate}
\item It ought to be that notified bodies fulfill the requirements in Article 31. 
\item It ought to be that if a notified body does not comply with the requirements in Article 31, further investigations are started. 
\item If a notified body does comply with the requirements in Article 31, no further investigation must be started. 
\item Concrete Situation: A notified body does not fulfill the requirements in Article 31. 
\end{enumerate}
\end{quote}
Reformulated as above, the example can be treated like a typical CTD. 

\subsection{Agency and Agentive Obligations}

Since different agents interact with an AI system during its lifetime, it is plausible that these agents have different duties towards the system based on their relation to it. Such obligations then are not general but agentive: They only hold for a specific type of agent, e.g. for providers or importers. For instance, the sentence 
\begin{quote}
"Providers of high-risk AI systems shall (...) have a quality management system in place which complies with Article 17" \cite[\S16(c)]{ai-act} 
\end{quote}
is such an agentive obligation because it expresses what the provider ought to do. Agentive obligations are different from general obligations, and must be distinguishable.

\subsection{Beliefs of Agents}

Apart from agentive obligations, another modality relating to agency is crucial in the AI Act: Agent beliefs. For example, Article 20 states: 

\begin{quote} 
"Providers of high-risk AI systems which consider or have reason to consider that a high-risk AI system that they have placed on the market or put into service is not in conformity with this Regulation shall immediately take the necessary corrective actions to bring that system into conformity, to withdraw it, to disable it, or to recall it, as appropriate." \cite[\S20(1)]{ai-act} 
\end{quote}

\indent Here, the provider's belief that a high-risk AI system is not in conformity with the regulation is relevant, not whether the system is, in fact, not conforming. Beliefs can be either right or wrong, and it is essential to differentiate them from facts. 

\subsection{Temporality and Temporal Notions}

Time is an important concept within the AI Act, with many articles specifying obligations that must be fulfilled before certain events occur. For instance, Article 23 states: 
\begin{quote}
"Before placing a high-risk AI system on the market, importers shall ensure that the system is in conformity with this Regulation by verifying that (...) "\cite[\S23(1)]{ai-act} 
\end{quote}

This obligation for importers is tied to a specific period and point in time; it must be fulfilled \textit{before} the system is placed on the market.

Additionally, some sentences contain hidden temporal notions, such as this phrase in Article 9: "When implementing the risk management system as provided for in paragraphs 1 to 7, providers shall…" \cite[\S9(9)]{ai-act}. The sentence implies simultaneity which becomes obvious if it is rephrased as follows: "While implementing the risk management system, providers are obligated to (...)." 

\subsection{Exceptions}

Within the AI Act, the concept of exceptions from general rules emerges as a recurring theme. Consider this sentence from Article~5: 
\begin{quote}
"The following AI practices shall be prohibited: (...) the use of ‘real-time’ remote biometric identification systems in publicly accessible spaces for the purposes of law enforcement, unless and in so far as such use is strictly necessary for one of the following objectives:...” \cite[\S5(1)]{ai-act}  
\end{quote}

This sentence outlines a general rule – the prohibition of certain AI practices – followed by exceptions based on specific objectives. 

\subsection{Fuzziness}

Fuzzy statements can be found in many places within the AI Act. They can be identified by their vagueness and usually involve statements that are not true or false, but true or false to a certain degree. The following expressions are examples of such fuzziness: 
\begin{quote}
"unless and in as far as such use is strictly necessary for one of the following objectives" \cite[\S5(1)]{ai-act}, "to the extent to which..." \cite[\S7]{ai-act}, and "at a minimum" \cite[\S11(1)]{ai-act}. 
\end{quote}
All these statements express vague notions or involve degrees of truth or necessity. 

\subsection{
Combinations of 
Modalities}

The modalities of the AI Act do not appear in isolation. Some of the examples discussed in Section 2.2 have already shown that two or more modalities are often combined within a single sentence. While this was ignored in the previous discussion for the sake of simplicity, it cannot be ignored when constructing a logical system intended to cover the entire AI Act, or larger coherent parts of it. Such a system must not only accurately represent all the relevant modalities individually, but also capture what happens when they are combined.

\section{Which Logics?}

 In this section we discuss logics that seem suitable to facilitate a representation of the AI Act in HOL.

 Standard Deontic Logic (SDL) is the most studied system of deontic logic. It contains a monadic deontic obligation operator depending on an accessibility relation and can successfully express and reason with obligations, permissions, and prohibitions \cite{sep-logic-deontic}. However, SDL reaches its limits and leads to inconsistencies in CTD scenarios \cite{dov_gabbay_handbook}.

 A logic that enables expressing and reasoning with CTDs is Dyadic Deontic Logic (DDL) by Carmo and Jones \cite{carmo_deontic_2002,CarmoJ22}. It differentiates between ideal and actual obligations and introduces a conditional obligation operator, thereby enabling the expression of CTDs that arise when ideal obligations have been violated. Related to DDL is Åqvist's system E for conditional obligation \cite{Aqvist2002}.
 
 To represent agentive obligations, a modal logic of agency is needed. While there are many ideas for formulating such a logic \cite{Anderson1962, Fitch1963, vonWright1963, von_wright_logic_1981, Kanger1972, brown_logic_1988}, the most prominent theory of agency is a branch called Seeing-To-It-That (STIT) logic that originated in the works of Belnap \cite{belnap_backwards_1991} and Belnap and Perloff \cite{Belnap1988-BELSTI}. STIT theory introduces a STIT operator of the form
$stit_a\ F$,
which expresses that an agent \textit{a} sees to it that \textit{F} holds. Given that agency appears in the AI Act mainly via agentive obligations, a suitable agency logic must be able to formulate ought-to-do statements for specific agents. Several authors have researched this issue and presented their approaches, which could be apt to help represent the AI Act \cite{van_berkel_neutral_2019, horty_belnap_1995, Horty2001, Murakami1998, Xu2015}.

Moving to the representation of agent beliefs, 
the field 
called epistemic logic becomes interesting. Epistemic logic allows for the exploration of different ideas of knowledge and beliefs of agents and their logical relations \cite{sep-logic-epistemic}.  Popular epistemic logics include modal logics S4 and S5 based on possible world semantics, and providing modal operators representing what agents \textit{know} and \textit{believe}.

For the representation of temporal constructs, we may turn to temporal logic, a class of modal logics that can formally represent and reason with temporality \cite{sep-logic-temporal}. A popular system is called Tense Logic (TL) and dates back to the work of Prior \cite{Prior1959, Prior1967, NPrior1968, Lejewski1}. TL treats propositions as tensed and introduces temporal operators into the language, thereby allowing for the representation of temporal relations like the ones in the AI Act.

Non-monotonic logic provides a way to deal with exceptions from general rules. It is designed to handle defeasible inference, enabling reasoners to retract conclusions when warranted by additional information, such as the presence of conditions allowing for an exception \cite{sep-logic-nonmonotonic}.

Finally, a logic that can do justice to vague and fuzzy statements is called fuzzy logic. Fuzzy logic assigns a degree of truth to a proposition that is expressed on a scale within the real unit interval [0, 1], with 0 equal to total falsehood and one equal to total truth~\cite{sep-logic-fuzzy}. From the point of view of LogiKEy, however, the fuzzy logic variants of H\'ajek \cite{Hajek98} are particularly interesting, since they can be seen as generalized multi-value logics and as such are more amenable to formalization; in fact, \cite{Hajek98} provides a good blue-print for doing so in LogiKEy extending the formalization idea presented in \cite{J33}.

\section{Initial Experiments}

In this section we give a brief summary of the findings and results from the experiments on embedding some of the logics using the LogiKEy approach and using these logics to represent parts of the AI Act; for details see \cite{ma}. Experiments have so far focused on only a selection of the identified modalities, namely obligations, prohibitions, permissions, CTDs, agency, and agentive obligations. 

 LogiKEy leverages higher-order logic (HOL) as a unifying meta-logic to enable the modelling of diverse object logics via shallow semantic embeddings \cite{J41}. This integration is facilitated, for example, within the higher-order proof assistant tool \textit{Isabelle/HOL}, which comes with state-of-the-art automated theorem provers (ATP), satisfiability modulo theories (SMT) solvers, and model finders; however, other proof assistant systems, such as Lean or Coq, or the TPTP infrastructure, could be employed instead. 

To represent obligations (as well as permissions and prohibitions), SDL was identified as a suitable logic in Section 3. A trustworthy embedding of SDL in Isabelle/HOL using the LogiKEy approach already exists \cite{SDLemb} and has been used to represent parts from the AI Act, particularly Article 5 on prohibited AI systems. This was successful: The paragraphs could be adequately represented, and both Sledgehammer (an automated theorem proving tool integrated with Isabelle/HOL) and Nitpick \cite{nitpick} (a model finder integrated with Isabelle/HOL) could reason with them correctly \cite{ma}.
Sledgehammer was used to automatically check the validity of example judgements, and Nitpick to find countermodels for invalid judgements and also to prove the consistency of whole contexts.

Like SDL, DDL already has a faithful embedding in Isabelle/HOL \cite{benzmüller_dyadic_2022}. Using this embedding it was possible to represent selected CTDs from the AI Act without problems. The formalization of the selected examples from the AI Act was straightforward, and Nitpick and Sledgehammer could work with them successfully \cite{ma}. Since DDL can express everything SDL can and more~\cite{dov_gabbay_handbook}, the DDL embedding is suitable for representing obligations, prohibitions, permissions, and CTDs. Encodings of Article 5 and selected CTDs from the AI Act in DDL can be found in the appendix.%
\footnote{The Isabelle/HOL source files can also be downloaded here: \url{http://logikey.org/tree/master/2025-ICAIL-Data}.} Further work includes similar experiments adapting an existing embedding of Åqvist's system E in Isabelle/HOL~\cite{J68}.

Additionally, logics for the representation of agency and  agentive obligations were explored. In particular, an embedding into HOL of Temporal Deontic STIT Logic \cite{van_berkel_neutral_2019}, a logic appropriate for representing both actions of agents and agentive obligations, was created. While most axioms postulated in \cite{van_berkel_neutral_2019} were provable via Sledgehammer and none were refuted, Nitpick could not find a model confirming the consistency of the embedding. This indicates that TDS logic has (presumably) only infinite models, making it unsuitable for model finding/checking in the context of any formalization of AI Act~\cite{ma} with existing tools.

As an alternative to TDS, the authors explored the possibility of extending DDL with additional operators for the representation of agency and agentive obligations. Two different variants were introduced: One representing agent types as constants, with separate accessibility relations and obligation operators for each agent, and one representing agent types as function types, with a general accessibility relation and a general obligation operator taking an agent as an input parameter. Both variants introduced a STIT-like operator as a constant and equipped it with minimal axiomatization, namely to ensure that anything an agent sees to actually holds  \cite{ma}.

While the first variant is more restrictive, with agentive obligations always applying to a generic agent type without the possibility of specifying further constraints or relations, the second variant is more flexible, allowing groupings based on characteristics of agent types or relations and formulating obligations for specific subsets of agents. Unfortunately, the second variant performs worse in other aspects: It struggles with the representation of CTDs, and Nitpick can not find a model for cardinalities of i (domain of possible worlds) bigger than 1. For the first variant, a model is found up to a cardinality of i=2, and CTDs can be represented without problems. For both variants, all but one of the prominent lemmas for DDL as studied by \citeauthor{benzmüller_dyadic_2022} could be proven \cite{ma}. Whether either of the variants can be adapted or simplified to perform better remains an open question. The embedding of TDS and the two extensions of DDL can be found in the appendix. 

The experiments have shown that while elementary concepts such as obligations and CTDs can be formalized and reasoned with correctly in HOL, the representation of complex constructs such as agency requires more sophisticated logics. Reasoning can be challenging for model finders and automated theorem provers, and further experiments need to be conducted to explore feasibility and automation performance as the studied contexts grow.

\section{Conclusion}
An extensive analysis of the AI Act was carried out, which led to the identification of different contained modalities that pose a challenge to attempts to formalise and reason with it: obligations, permissions, prohibitions, CTDs, agency and agentive obligations, beliefs of agents, temporality and temporal notions, fuzziness, and exceptions. Suggestions for logics suitable to represent these modalities were presented, and results of some first experiments with these logics and examples from the AI Act were reported. As of now, experiments have only been conducted for the following modalities (and some combinations of) : obligations, permissions, prohibitions, CTDs, agency, and agentive obligations.

While the presented project is not yet completed, the reported material provides a starting point and references for future research.
In the LogiKEy context, the challenge is to create embeddings of the suggested logics, choose and provide suitable logic combinations, and experiment with increasingly larger parts of the AI Act.

Based on the information gathered here, the possibility of representing legal information in HOL and using it in automated reasoning should be further explored and improved.

\paragraph*{Acknowledgements}
We are grateful to the anonymous reviewers for their helpful feedback and comments.

\bibliographystyle{ACM-Reference-Format}
\bibliography{bibliography}

\newpage
\onecolumn
\section{Appendix}

\subsection{Article 5 in DDL}

\begin{figure}[h]
    \centering
    \includegraphics[width=1 \linewidth]{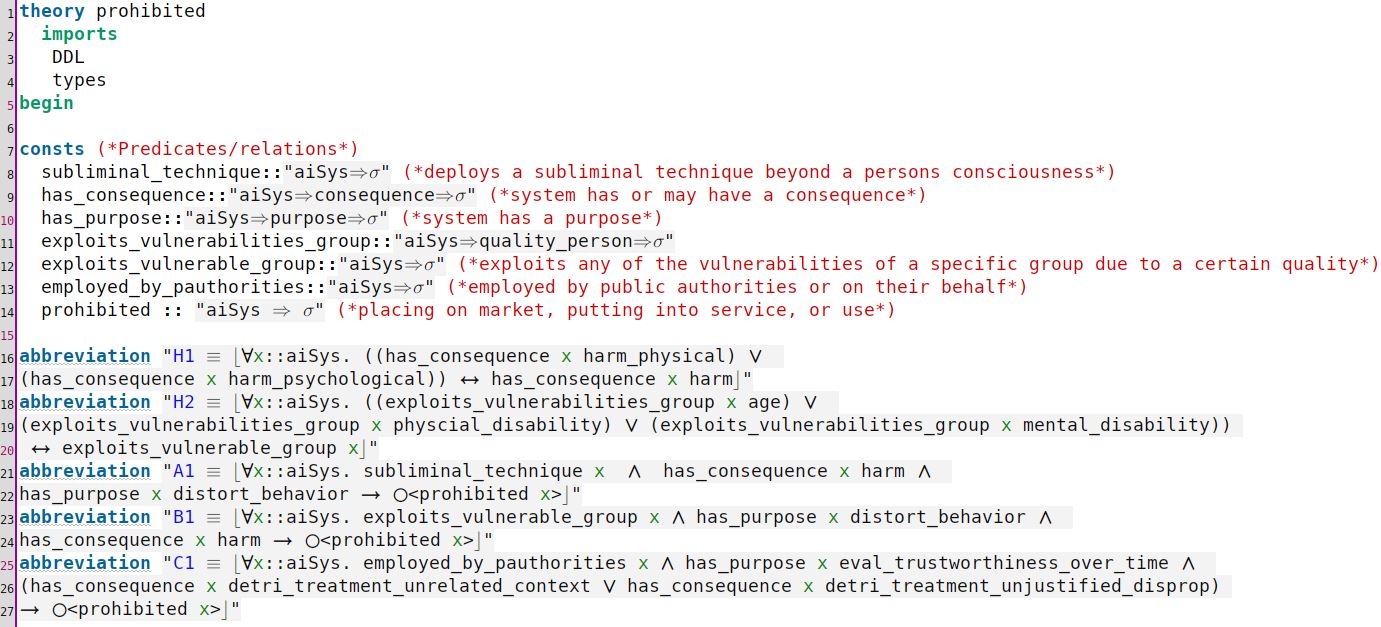}
    \caption{Article 5 in DDL, Part 1}
    \Description{Article 5 in DDL, Part 1}
    \label{art5:1}
\end{figure}

The following screenshots illustrate the representation of the following part of Article 5  in Isabelle HOL:
\begin{quote}
    "1. The following AI practices shall be prohibited:\\
    (a) the placing on the market, the putting into service or the use of an AI system that deploys subliminal techniques beyond
    a person’s consciousness or purposefully manipulative or deceptive techniques, with the objective, or the effect of
    materially distorting the behaviour of a person or a group of persons by appreciably impairing their ability to make an
    informed decision, thereby causing them to take a decision that they would not have otherwise taken in a manner that
    causes or is reasonably likely to cause that person, another person or group of persons significant harm;\\
    (b) the placing on the market, the putting into service or the use of an AI system that exploits any of the vulnerabilities of
    a natural person or a specific group of persons due to their age, disability or a specific social or economic situation, with
    the objective, or the effect, of materially distorting the behaviour of that person or a person belonging to that group in
    a manner that causes or is reasonably likely to cause that person or another person significant harm;\\
    (c) the placing on the market, the putting into service or the use of AI systems for the evaluation or classification of natural
    persons or groups of persons over a certain period of time based on their social behaviour or known, inferred or
    predicted personal or personality characteristics, with the social score leading to either or both of the following:\\
    (i) detrimental or unfavourable treatment of certain natural persons or groups of persons in social contexts that are
    unrelated to the contexts in which the data was originally generated or collected;\\
    (ii) detrimental or unfavourable treatment of certain natural persons or groups of persons that is unjustified or
    disproportionate to their social behaviour or its gravity;" \cite[\S5(1)]{ai-act}.
\end{quote}
As hinted at in Figure \ref{art5:1} in the import section, the embedding of DDL in Isabelle/HOL from \cite{benzmüller_dyadic_2022} is reused here. 

Lines 7-14 define constants, which are then used in lines 16-27 to define the rules given in paragraph 1, a-c. Lines 29-42 construct an example case and test two lemmas. The first constructs a conclusion that should hold given the rules and the properties of the example system. As expected, it can be proven using the Sledgehammer tool. On the other hand, the second one is missing part of the information from the previous lemma and should not hold. Nitpick finds a counterexample.  

\begin{figure}[h]
    \centering
    \includegraphics[width=1\linewidth]{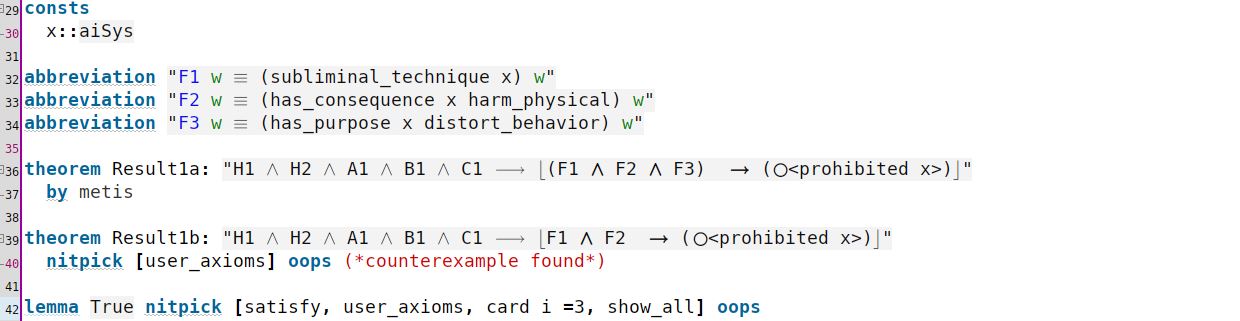}
    \caption{Article 5 in DDL, Part 2}
    \Description{Article 5 in DDL, Part 2}
    \label{art5:2}
\end{figure}

Note how the formalization of the theorem statements in Figure \ref{art5:2} distinguishes between global (H1-C1) and local (F1-F3) assumptions.

\newpage

\subsection{Selected CTD Examples in DDL}
\begin{figure}[h]
    \centering
    \includegraphics[width=1\linewidth]{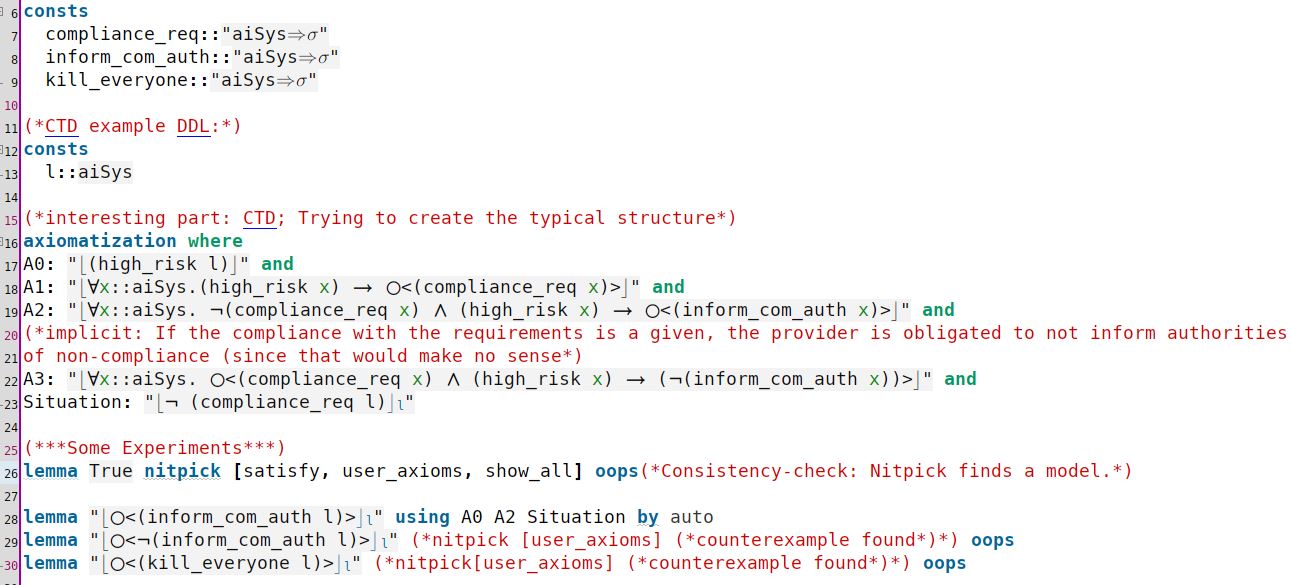}
    \caption{Article 20, CTD Example in DDL}
    \Description{Article 20, CTD Example in DDL}
    \label{art20:1}
\end{figure}

\begin{figure}[h]
    \centering
    \includegraphics[width=1\linewidth]{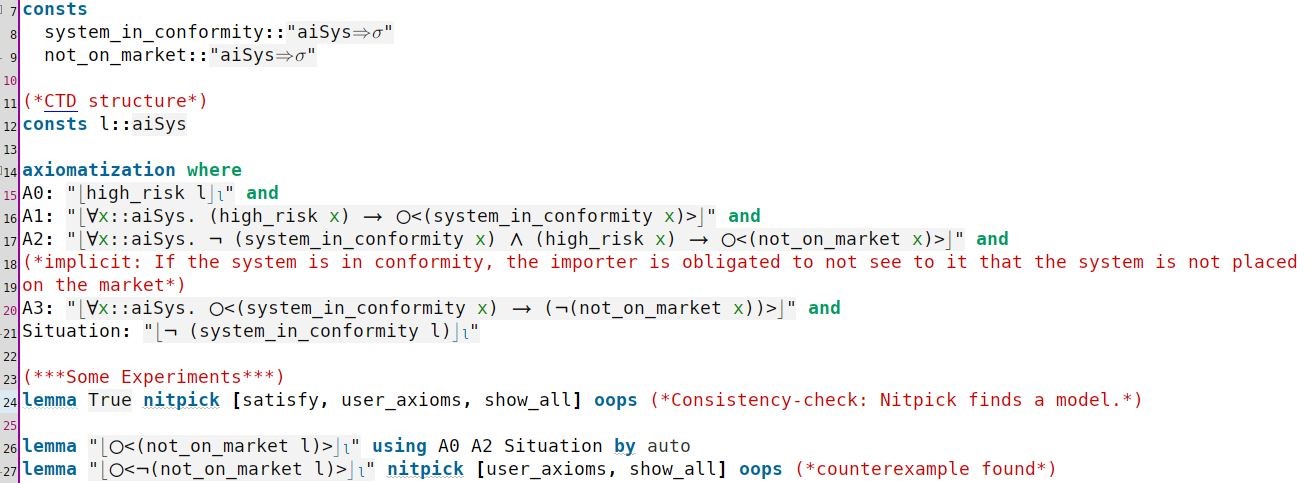}
    \caption{Article 24, CTD Example in DDL}
    \Description{Article 24, CTD Example in DDL}
    \label{art24:1}
\end{figure}

Figures \ref{art20:1} and \ref{art24:1} show representations of two CTDs. The first arises from a combination of Articles 16 and 20:
\begin{quote}
    "Providers of high-risk AI systems shall:\\
(a) ensure that their high-risk AI systems are compliant with the requirements set out in Section 2;
\cite[\S16]{ai-act}
\end{quote}
\begin{quote}
    1. Providers of high-risk AI systems which consider or have reason to consider that a high-risk AI system that they have
    placed on the market or put into service is not in conformity with this Regulation shall immediately take the necessary
    corrective actions to bring that system into conformity, to withdraw it, to disable it, or to recall it, as appropriate. They shall
    inform the distributors of the high-risk AI system concerned and, where applicable, the deployers, the authorised
    representative and importers accordingly."
\cite[\S20]{ai-act}
\end{quote}

The same excerpt from Article 16 results in another CTD when combined with the following part from Article 24:

\begin{quote}
    "2. Where an importer has sufficient reason to consider that a high-risk AI system is not in conformity with this
    Regulation, or is falsified, or accompanied by falsified documentation, it shall not place the system on the market until it has been brought into conformity." \cite[\S24]{ai-act}
\end{quote}
Again, the embedding of DDL is used \cite{benzmüller_dyadic_2022}. 

In both cases, consistency is proven and Isabelle can successfully reason in the CTD structure, with Sledgehammer proving the desired lemmas and Nitpick providing counterexamples wrong statements. 


\subsection{TDS Embedding}

\begin{figure}[H]
    \centering
    \includegraphics[width=1\linewidth]{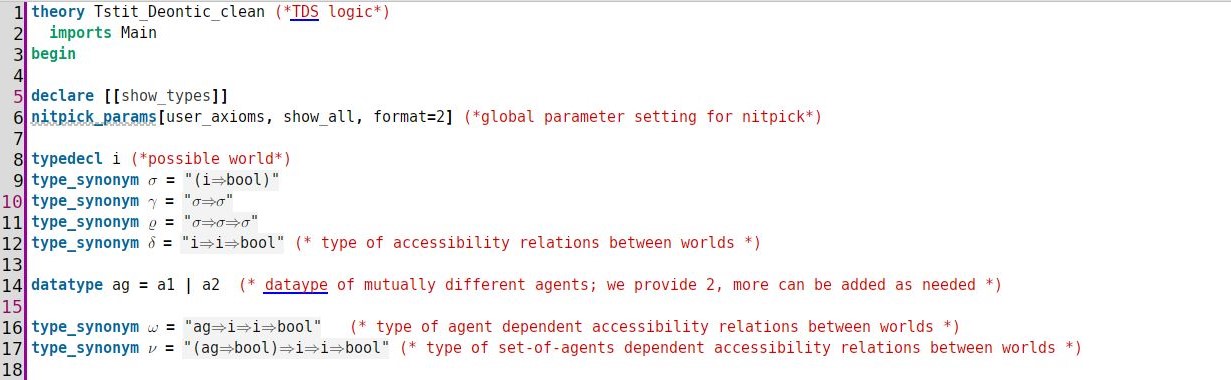}
    \caption{TDS Embedding Part 1}
    \Description{TDS Embedding Part 1}
    \label{tds1}
\end{figure}

\begin{figure}[H]
    \centering
    \includegraphics[width=1\linewidth]{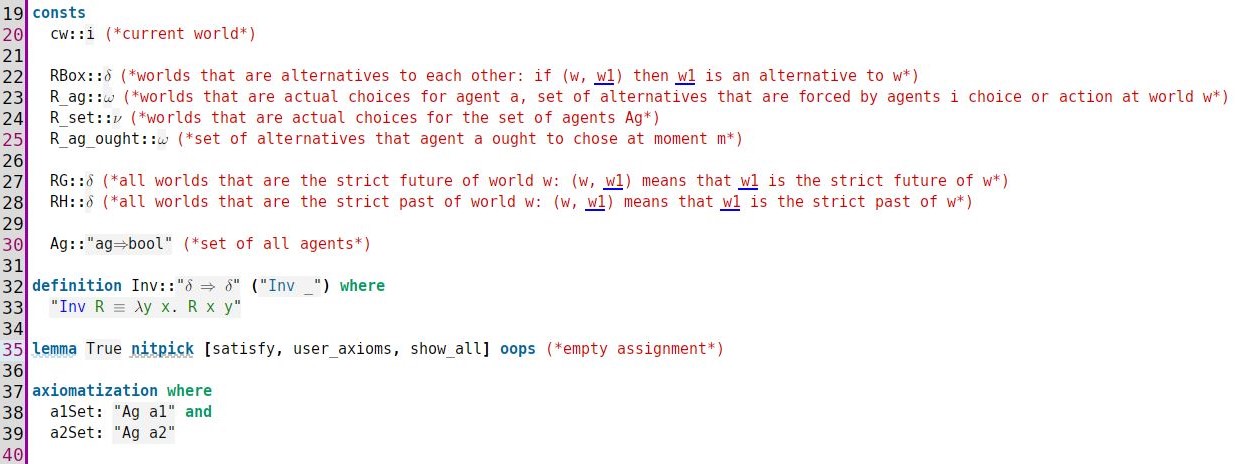}
    \caption{TDS Embedding Part 2}
    \Description{TDS Embedding Part 2}
    \label{tds2}
\end{figure}

\begin{figure}[H]
    \centering
    \includegraphics[width=1\linewidth]{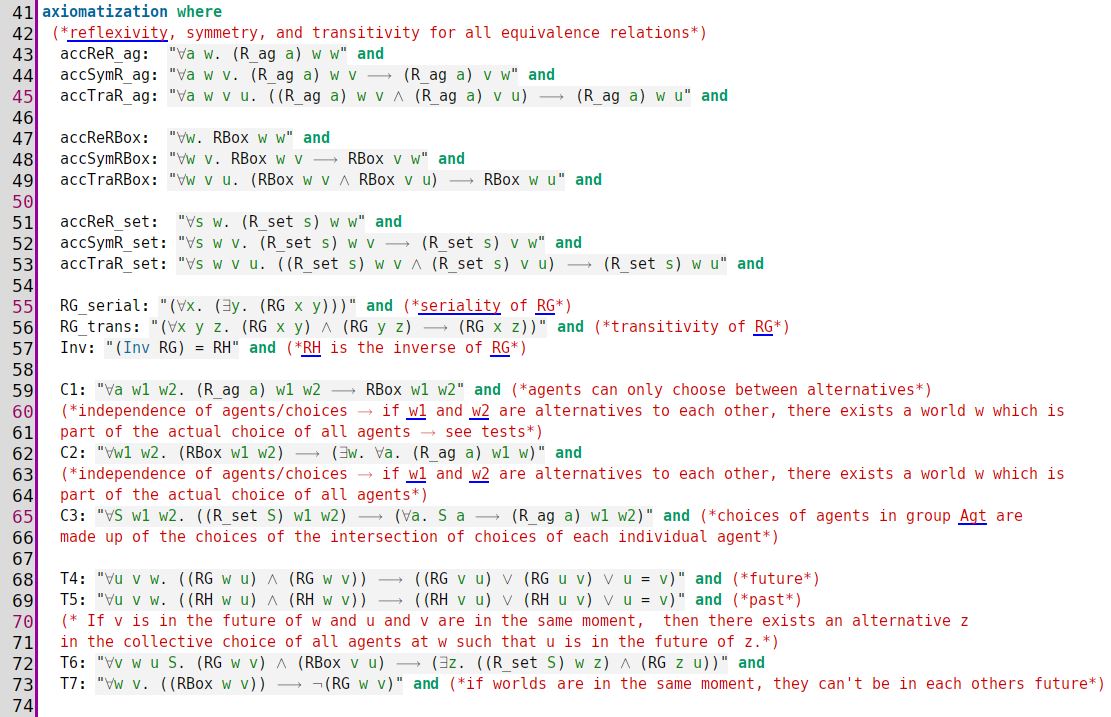}
    \caption{TDS Embedding Part 3}
    \Description{TDS Embedding Part 3}
    \label{tds3}
\end{figure}

\begin{figure}[H]
    \centering
    \includegraphics[width=1\linewidth]{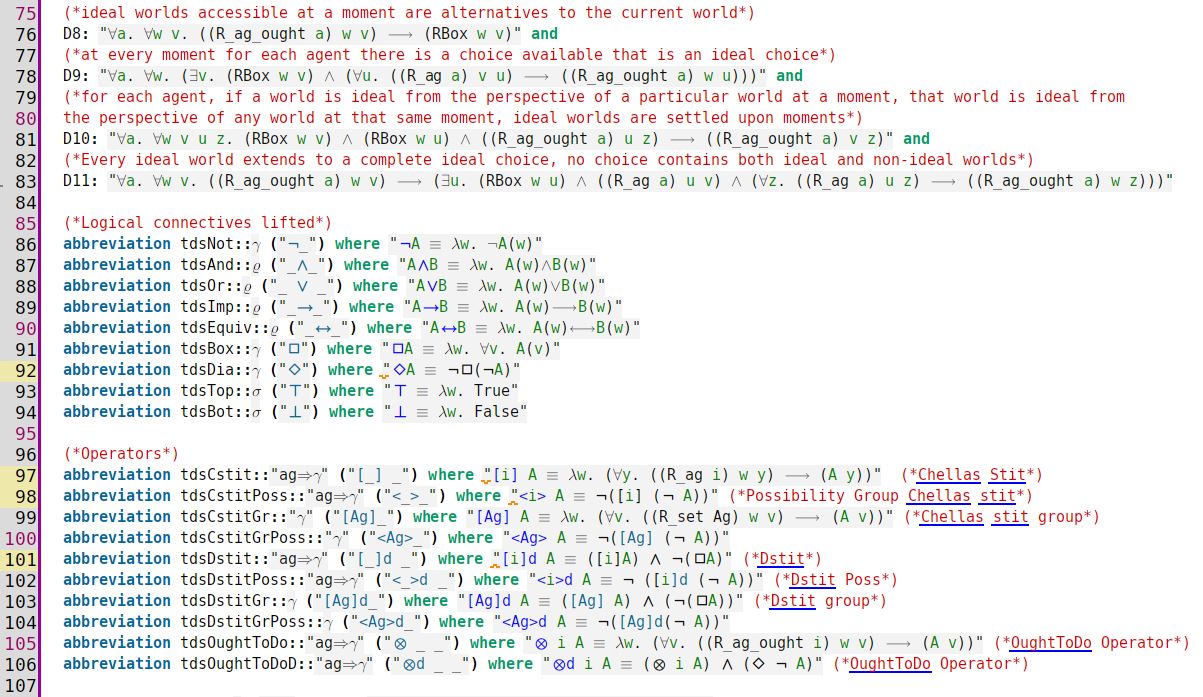}
    \caption{TDS Embedding Part 4}
    \Description{TDS Embedding Part 4}
    \label{tds4}
\end{figure}

\begin{figure}[H]
    \centering
    \includegraphics[width=1\linewidth]{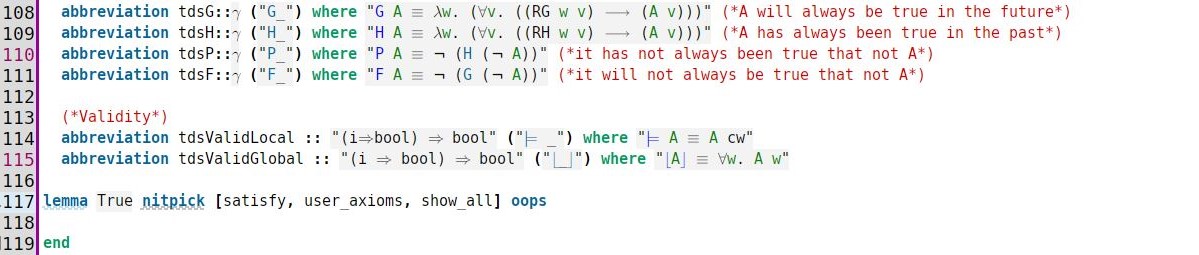}
    \caption{TDS Embedding Part 5}
    \Description{TDS Embedding Part 5}
    \label{tds5}
\end{figure}

Figures \ref{tds1}, \ref{tds2}, \ref{tds3}, \ref{tds4}, and \ref{tds5} show the embedding of TDS that has been created by the first author. 

\begin{figure}[H]
    \centering
    \includegraphics[width=1\linewidth]{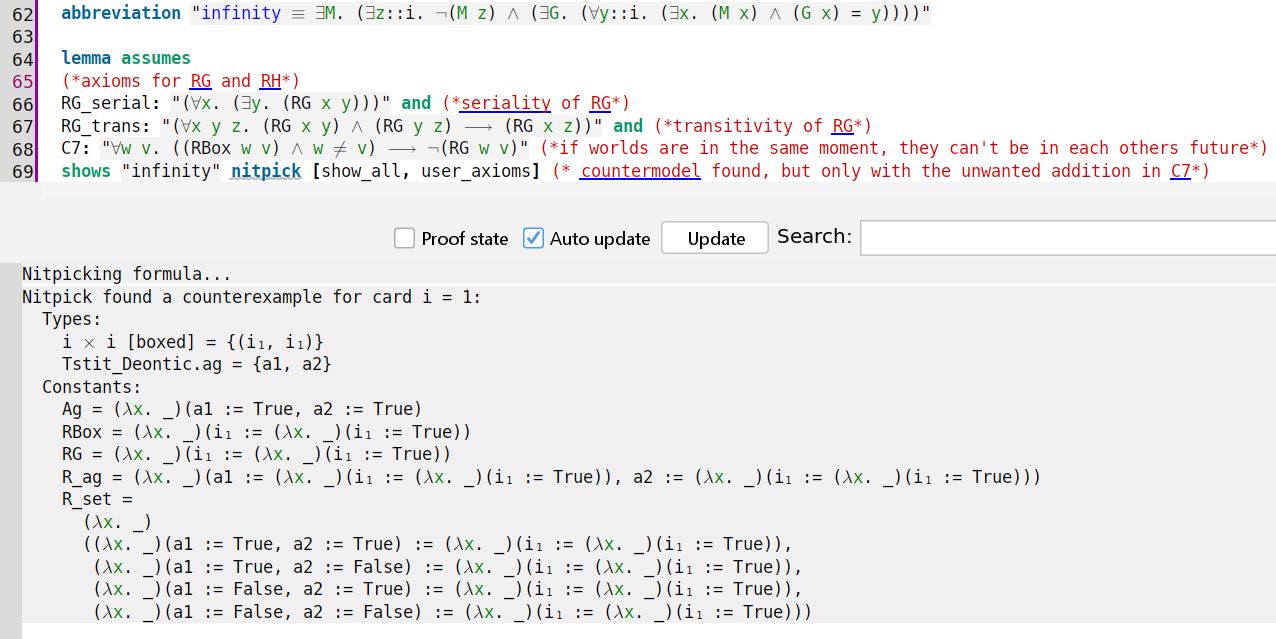}
    \caption{TDS Infinity Proof Part 1}
    \Description{TDS Infinity Proof Part 1}
    \label{tds6}
\end{figure}

\begin{figure}[H]
    \centering
    \includegraphics[width=1\linewidth]{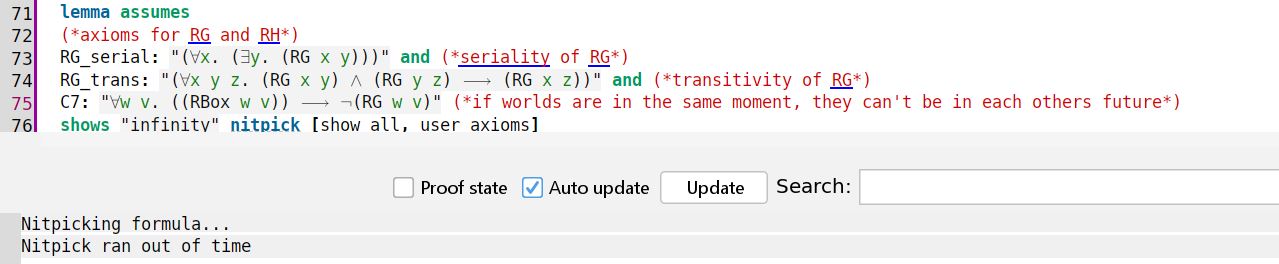}
    \caption{TDS Infinity Proof Part 2}
    \Description{TDS Infinity Proof Part 2}
    \label{tds7}
\end{figure}

Lines 8-17 define the types of worlds, agents, and the accessibility relations. The constants defined in lines 19-30 represent the current world, accessibility relations of the logic, and the set of agents. In the axiomatization, the first author defines the set of agents as containing the two agents a1 and a2, specify reflexivity, symmetry, and transitivity for all equivalence relations, seriality, transitivity, and the inverse relation for the future relation RG, and the constraints imposed on the different relations (lines 37-83). Afterward, lines 85-94 define the lifted connectives, whereas lines 96-111 define the operators of TDS logic. Lines 113-115 define notions of validity. Finally, line 117 calls Nitpick to find a model of the embedding. As explained in Section 4, this does not succeed. The infinity proof is visible in Figures \ref{tds6} and \ref{tds7}.

For a complete explanation of the TDS embedding and the infinity proof, please refer to \cite{ma}. 

\newpage

\subsection{Extended DDL First Variant}

\begin{figure}[h]
    \centering
    \includegraphics[width=1\linewidth]{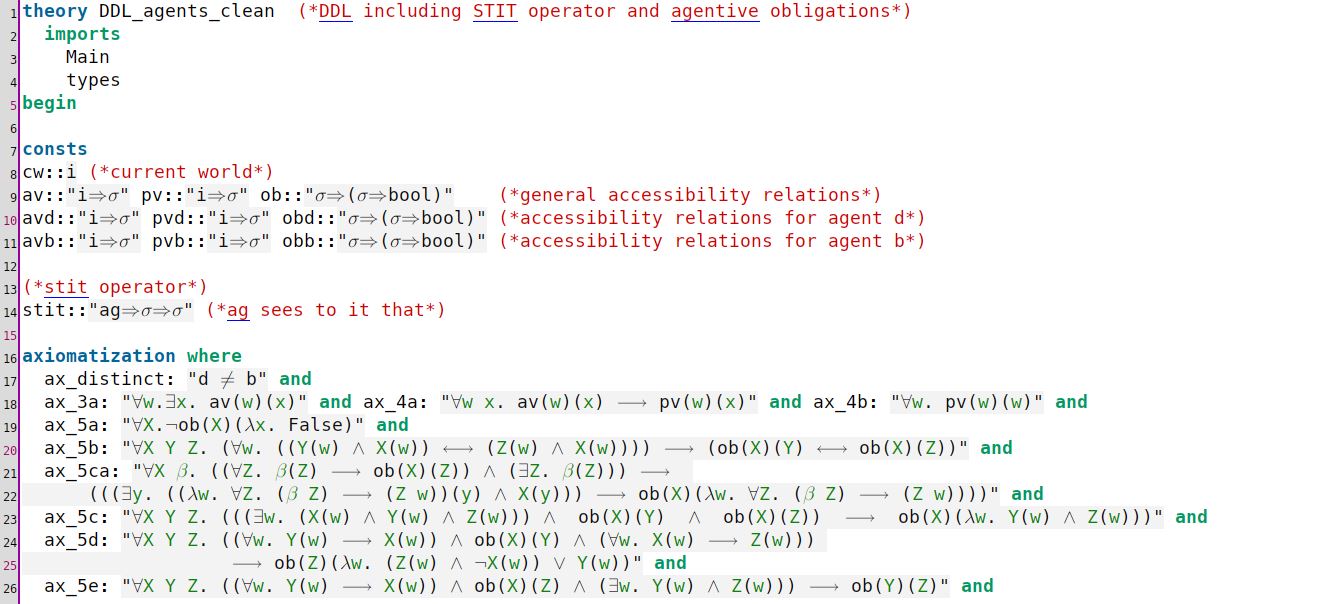}
    \caption{Extended DDL First Variant Part 1}
    \Description{Extended DDL First Variant Part 1}
    \label{DDL1_1}
\end{figure}

\begin{figure}[h]
    \centering
    \includegraphics[width=1\linewidth]{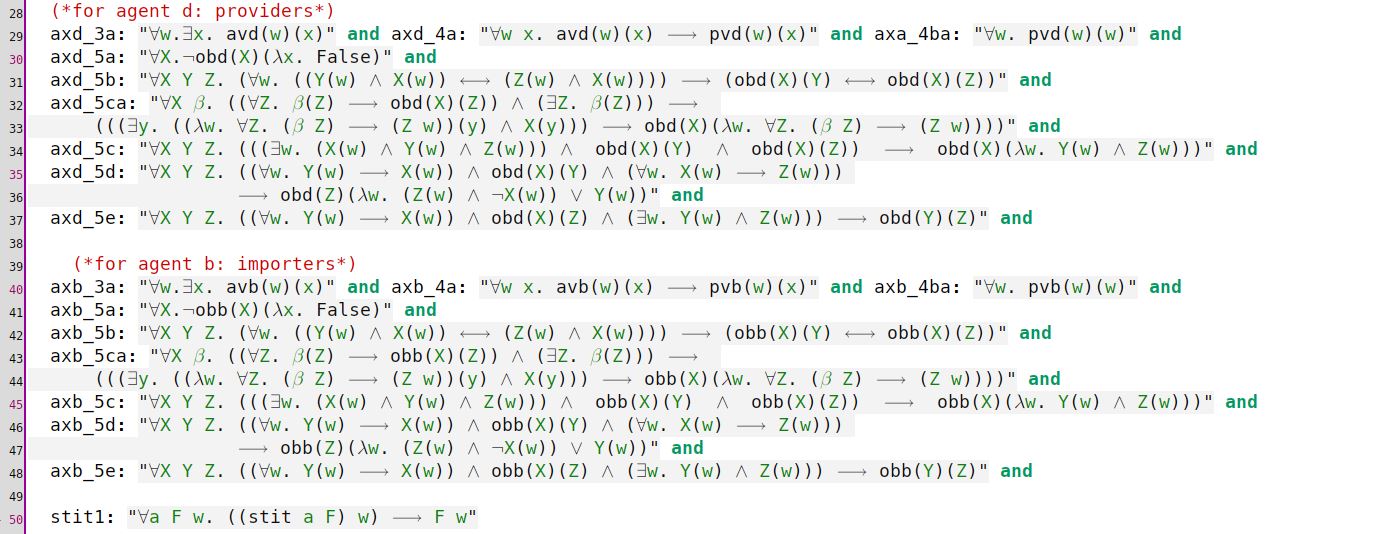}
    \caption{Extended DDL First Variant Part 2}
    \Description{Extended DDL First Variant Part 2}
    \label{DDL1_2}
\end{figure}

\begin{figure}[h]
    \centering
    \includegraphics[width=1\linewidth]{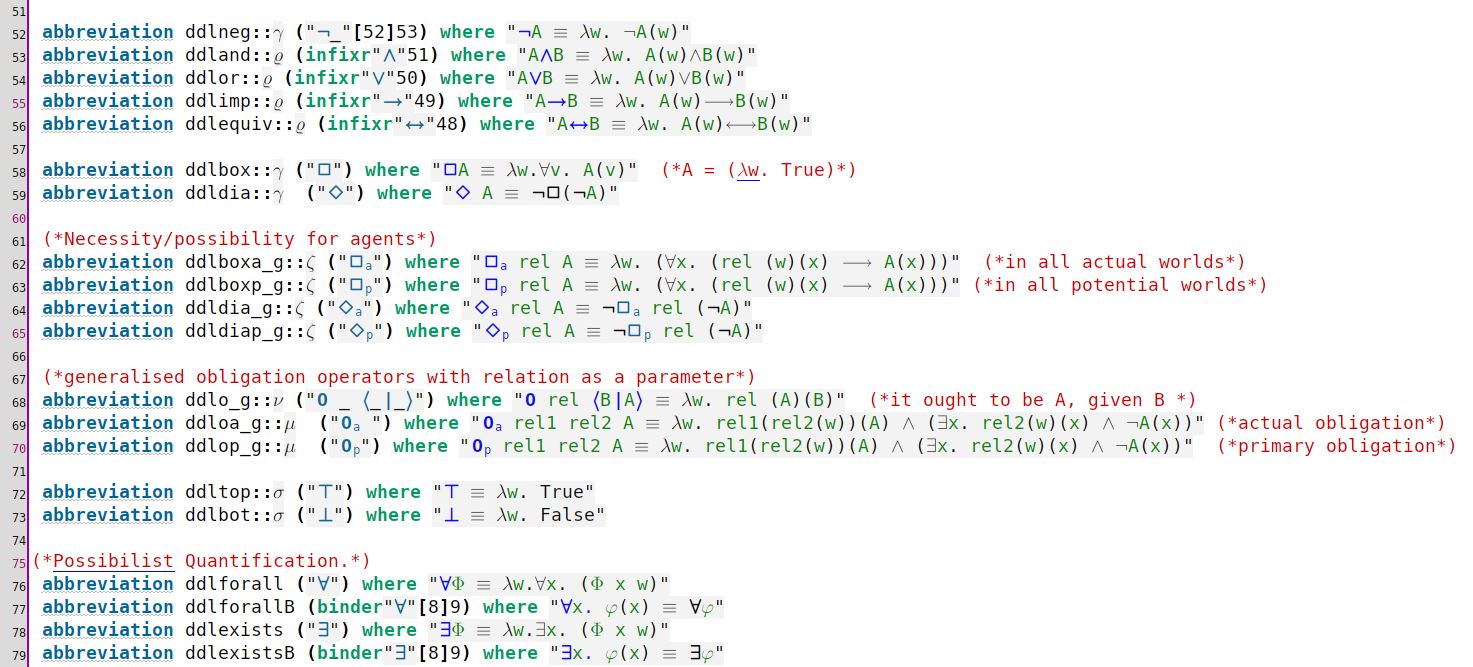}
    \caption{Extended DDL First Variant Part 3}
    \Description{Extended DDL First Variant Part 3}
    \label{DDL1_3}
\end{figure}

\begin{figure}[h]
    \centering
    \includegraphics[width=1\linewidth]{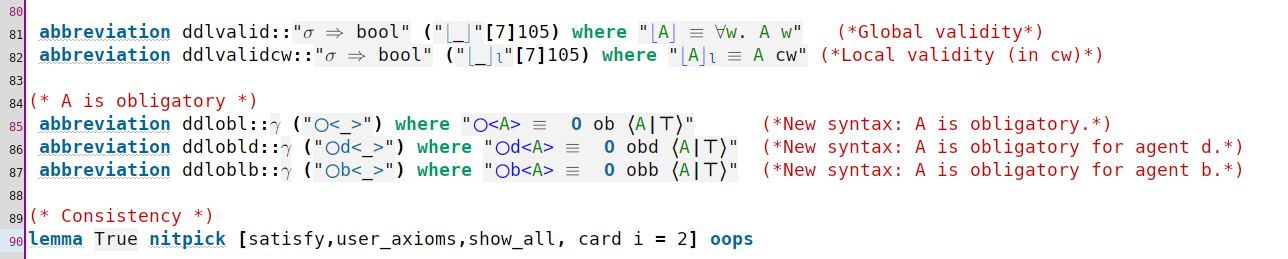}
    \caption{Extended DDL First Variant Part 4}
    \Description{Extended DDL First Variant Part 4}
    \label{DDL1_4}
\end{figure}

Figures \ref{DDL1_1}, \ref{DDL1_2}, \ref{DDL1_3}, and \ref{DDL1_4} show the first variant of Extended DDL created by the first author \cite{ma}. 

In contrast to the usual DDL embedding, this version uses two additional sets of accessibility relations, which relate to two different agents d and b (lines 10-11). They are axiomatized just like the general accessibility relations of DDL (lines 28-48). These are then used to define generalized obligation operators which take relations (the general one or the ones for the agents) as inputs (lines 67-70). Shortcuts for the general obligation operator and the two agentive obligation operators are declared in lines 84-87. 

Additionally, a constant representing the stit operator is introduced in line 14 and axiomatized in line 50. 

A more detailed explanation can be found in \cite{ma}. 

\newpage
\subsection{Extended DDL Second Variant}

\begin{figure}[H]
    \centering
    \includegraphics[width=1\linewidth]{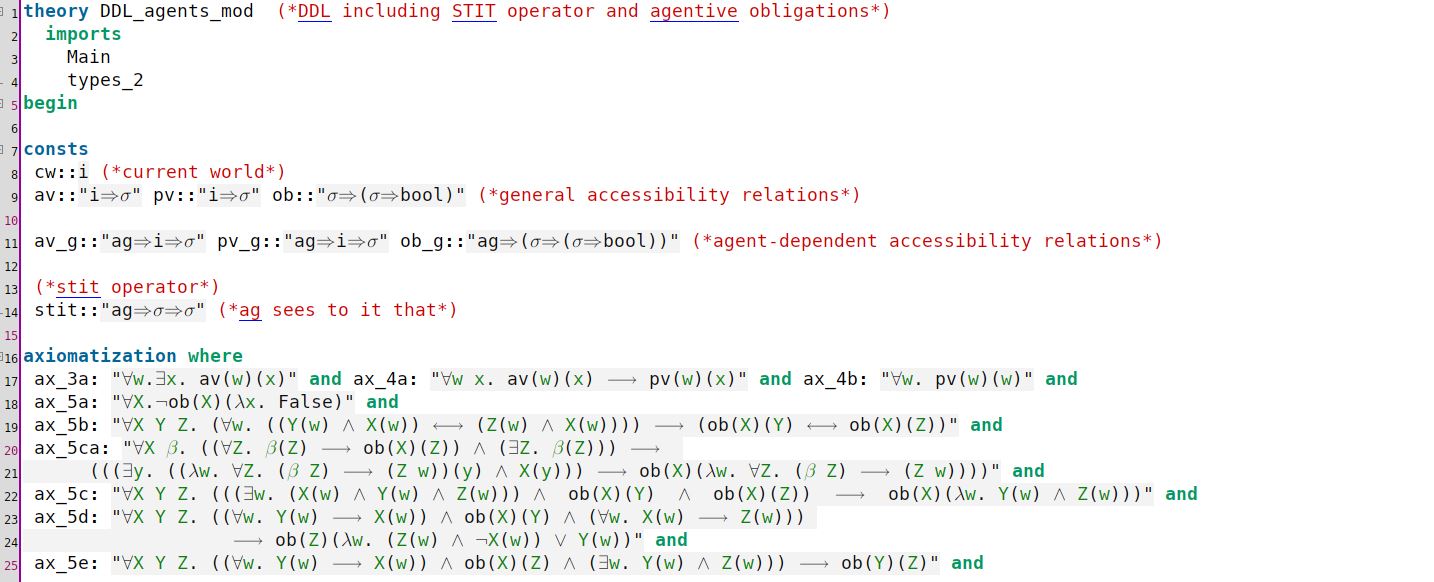}
    \caption{Extended DDL Second Variant Part 1}
    \Description{Extended DDL Second Variant Part 1}
    \label{DDL2_1}
\end{figure}

\begin{figure}[h]
    \centering
    \includegraphics[width=1\linewidth]{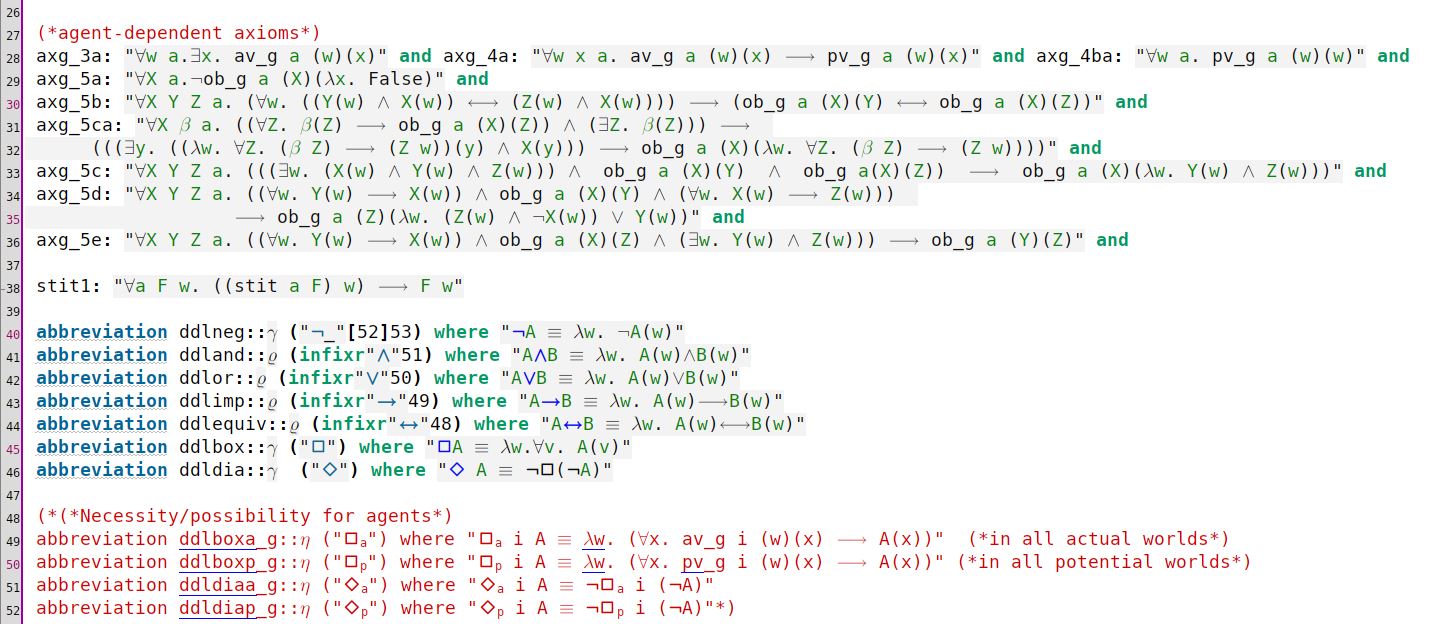}
    \caption{Extended DDL Second Variant Part 2}
    \Description{Extended DDL Second Variant Part 2}
    \label{DDL2_2}
\end{figure}

\begin{figure}[h]
    \centering
    \includegraphics[width=1\linewidth]{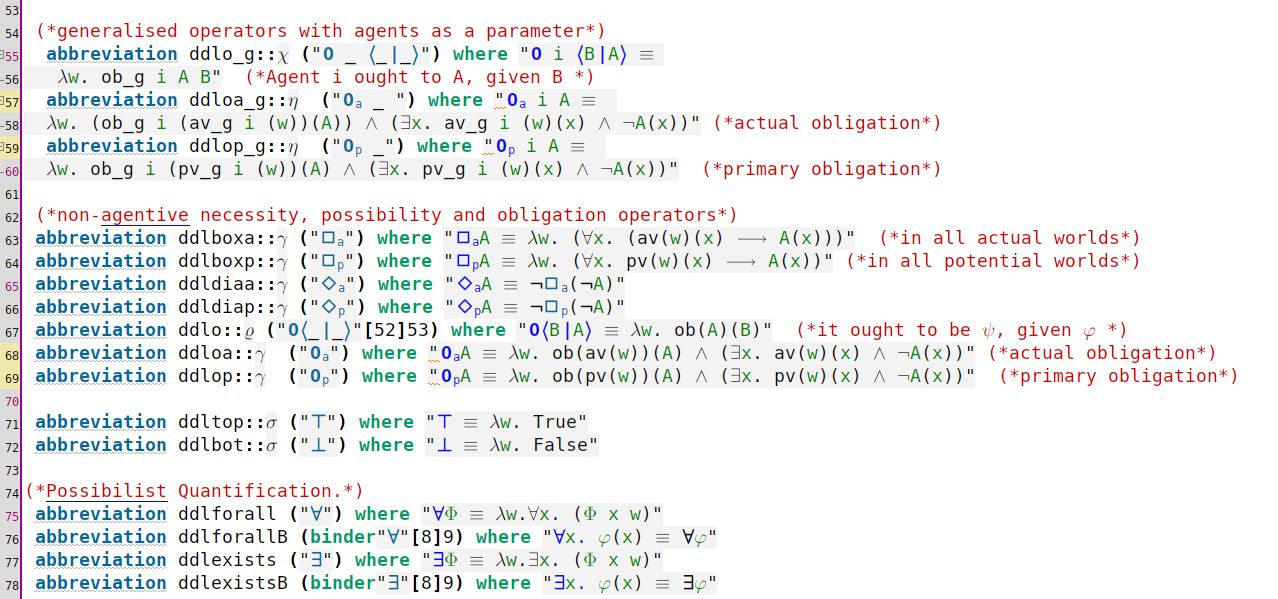}
    \caption{Extended DDL Second Variant Part 3}
    \Description{Extended DDL Second Variant Part 3}
    \label{DDL2_3}
\end{figure}

\begin{figure}[h]
    \centering
    \includegraphics[width=1\linewidth]{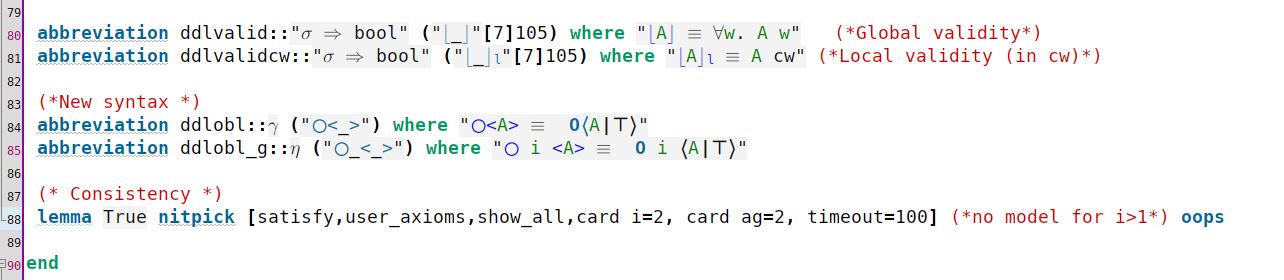}
    \caption{Extended DDL Second Variant Part 4}
    \Description{Extended DDL Second Variant Part 4}
    \label{DDL2_4}
\end{figure}

Finally, Figures \ref{DDL2_1}, \ref{DDL2_2}, \ref{DDL2_3}, and \ref{DDL2_4} show the second variant of Extended DDL created by the first author \cite{ma}.

Whereas the first variant introduced one set of accessibility relations for each additional agent, the second one works with a single, generalized set of accessibility relations, with the relations taking an agent as an input parameter (line 11). The axiomatization for this new set of accessibility relations can be found in lines 27-36. Using the new generalized accessibility relations, generalized obligation operators are defined in lines 54-60. They also take an agent as an input parameter. Abbreviations for the non-agentive and agentive accessibility relations are introduced in lines 83-85.

The design of the stit operator, including its axiomatization, is equivalent to the first variant.

For a more detailed explanation, please refer to \cite{ma}. 

\end{document}